# Depth-Aware Sensitivity Analysis of Mixture-of-Experts Models via Magnitude-Based Expert Masking

**Authors: Pradeep Kumar Sharma, Shantanu Godbole, Hritvik Shrivastava**

**Affiliation: Persistent Systems**

## Abstract

Mixture-of-Experts (MoE) architectures scale large language models (LLMs) while preserving computational efficiency through sparse activation. Despite their widespread adoption, the relative importance of individual MoE layers remains insufficiently characterized, particularly for model compression. This paper presents a systematic layer-wise sensitivity analysis of the Qwen3.6-35B-A3B model (40 MoE layers, 256 experts per layer, top-8 routing) using magnitude-based expert masking on the XLCoST cross-lingual code translation benchmark.

We conduct a multi-phase study spanning 100, 300, and 500 prompt evaluation scales across three H100 GPU servers. Our central finding is that **layer sensitivity is strongly depth-dependent**: early layers (0-9) and middle layers (10-29) are highly fragile to expert masking, while late layers (30-39), and especially very-late layers (35-39), tolerate aggressive masking of low-magnitude experts. Flat all-layer masking at 30% retains only 150/300 Good+Similar outputs at 300-prompt scale, whereas late-focused policies retain 249-255/300 while masking 640-1,145 experts. On a later 500-prompt held-out validation slice, the narrow very-late policy (layers 35-39 @ 50%) achieves the strongest quality/masked-expert tradeoff among tested candidates, retaining 419/500 Good+Similar outputs while masking only 640 of 10,240 total experts.

We additionally characterize top-k routing width reduction from 8 to 6 active experts per token, which shows a large observed wall-clock reduction on a 100-prompt probe with no Good+Similar loss, though it does not yet compose cleanly with aggressive expert masking. These findings provide an empirical foundation for depth-aware MoE expert masking and establish a practical path toward physical weight surgery, activation-based expert scoring, and training-based recovery.

## 1. Introduction

Rapid growth in large language model (LLM) scale has been driven by demand for high performance across language understanding and code generation tasks. As parameter counts have grown into the tens and hundreds of billions, Mixture-of-Experts (MoE) architectures have emerged as an effective approach to scaling model capacity without proportionally increasing per-inference cost. MoE models route each token through a gating mechanism that selects a small subset of expert feed-forward networks, enabling substantially larger parameter budgets at inference costs comparable to smaller dense models [1, 2].

Despite this success, fundamental questions about the **relative importance of individual MoE layers** remain open: Do all MoE layers contribute equally to downstream behavior? Are some depth regions more redundant than others? Can aggressive expert removal be applied selectively without degrading output quality? These questions have direct implications for compression, deployment efficiency, and future architecture design.

Existing pruning literature focuses largely on dense models [3, 4] or globally uniform sparsity schedules [5]. Work specifically characterizing layer-wise sensitivity in large-scale MoE systems is limited. Most prior MoE compression research concentrates on expert-level routing frequency, expert skipping, or global pruning budgets [6, 7], without large-scale empirical evaluation of how sensitivity varies systematically with layer depth.

This paper addresses this gap through a multi-phase empirical study of Qwen3.6-35B-A3B. Our main contributions are:

1. **Depth-dependent sensitivity.** We demonstrate empirically that early layers (0-9) and middle layers (10-29) are fragile under expert masking, while late layers (30-39) are substantially more tolerant.
2. **Rejection of uniform pruning.** At 300-prompt scale, flat 30% all-layer masking retains only 150/300 Good+Similar outputs, while late-focused schedules retain up to 255/300.
3. **Late-layer quality/masked-expert frontier.** Structured late-ramp, global late-budget, and very-late-only policies form a Pareto frontier. The 300-prompt discovery sweep favors the late ramp, while the 500-prompt held-out validation favors a narrower very-late-only policy.
4. **Routing width reduction.** Reducing active experts from 8 to 6 per token shows a large wall-clock reduction with no Good+Similar loss on a 100-prompt probe, but interacts non-additively with aggressive expert masking.
5. **Auxiliary thinking-phase expert isolation.** We include a phase-separated routing analysis on a related Qwen3.5-35B-A3B model that isolates prompt, thinking, and answer-phase expert usage. This is not used as causal evidence for the Qwen3.6 masking frontier, but it motivates phase-aware expert scoring.
6. **Reproducible experimental framework.** We provide a complete methodology including disjoint masking buckets, router-logit exclusion, mechanical integrity checks, baseline-relative generation scoring, and multi-server execution.

## 2. Background and Related Work

### 2.1 Mixture-of-Experts Architectures

The Mixture-of-Experts paradigm has roots in Jacobs et al. [8] and Jordan & Jacobs [9], who introduced the decomposition of complex problems into specialized subnetworks. In modern transformer-based LLMs, MoE layers replace dense feed-forward sub-layers with parallel expert networks and a learnable routing mechanism.

Fedus et al. [1] demonstrated scalability with Switch Transformer, routing each token to exactly one expert per layer. Jiang et al. [2] extended this to top-k routing with Mixtral 8x7B. DeepSeek-MoE [10] introduced finer-grained expert decomposition, while the Qwen-MoE family [11], including the Qwen3.6-35B-A3B model used here, employs 256 experts per layer with top-8 routing.

A key property relevant to this work is that routing decisions are made independently at each layer. This creates the possibility that different depth regions develop different redundancy profiles, which is the central hypothesis tested here.

### 2.2 Model Compression and Pruning

Sensitivity analysis quantifies how parameter changes affect model behavior. Optimal Brain Damage [3] and Optimal Brain Surgeon [12] established that parameters with small second-order loss contributions can be removed with minimal degradation. Magnitude-based pruning [13] operates under the simpler assumption that small-magnitude parameters contribute less to predictions. Despite its simplicity, magnitude pruning has proven effective, particularly when combined with iterative pruning and fine-tuning [4]. Han et al. [5] demonstrated large compression ratios through unstructured magnitude pruning.

### 2.3 MoE-Specific Compression

In the MoE setting, pruning carries additional complexity because expert use is input-dependent. Lu et al. [6] studied expert importance heterogeneity, showing that not all experts contribute equally. He et al. [7] analyzed structured pruning strategies for MoE models. Muzio et al. [14] explored expert redundancy and merging as an alternative to deletion. Zoph et al. [15] investigated expert capacity and routing efficiency in ST-MoE. These approaches motivate MoE compression, but they generally do not isolate depth as the main experimental variable.

### 2.4 Evaluation Task

We evaluate masking sensitivity on a structured code generation task drawn from the XLCoST benchmark [16], which provides aligned code snippets across multiple programming languages. Code translation is well-suited for sensitivity analysis because it requires the model to preserve algorithmic semantics across a non-trivial transformation, producing outputs whose structural validity and semantic fidelity can be assessed programmatically. The test split provides a sufficiently large pool to draw fixed, deterministic subsets for multiple experimental phases.

## 3. Experimental Setup

### 3.1 Model

All experiments use Qwen3.6-35B-A3B [11], containing 40 MoE layers with 256 experts per layer and top-8 routing, yielding approximately 3.3B active parameters per forward pass out of 35B total. Experiments use bfloat16 precision on NVIDIA H100 80GB HBM3 GPUs, with each run isolated to a dedicated two-GPU pair. Software: Hugging Face Transformers 5.8.0.dev0, PyTorch 2.10.0+cu128, Accelerate 1.13.0.

### 3.2 Dataset and Evaluation

The evaluation pool is drawn from the XLCoST benchmark [16]. Fixed deterministic subsets are drawn for each experimental phase:

| Phase | Prompt count | Purpose |
|---|---|---|
| Early smoke tests | 10 | Masking and decoding validation |
| Regional ablations | 100 | Fast sensitivity isolation |
| Discovery sweep | 300 | Policy search and ranking |
| Held-out validation | 500 | Larger-scale validation of frontier policies |

Each sample is formatted as a translation instruction requesting only the target-language output.

We employ two evaluation approaches:

1. **Teacher-forced token-level accuracy** (early experiments): The prompt and reference output are concatenated in a single forward pass; the proportion of correctly predicted reference tokens measures translation quality.

2. **Generation-based baseline-relative scoring** (primary metric): The model generates a free-form output (max_new_tokens = 1200). Each masked output is scored against the unmasked baseline generation using normalized code similarity and assigned to one of three categories:
   - **Good (G):** Normalized code exact match and masked output is syntactically valid.
   - **Similar (S):** Either exact-match invalid on both sides, or masked output is syntactically valid and normalized code similarity to baseline is at least 0.85.
   - **Bad (B):** Syntactically invalid output or material code divergence.

The combined **G + S** count is the primary quality retention metric. The metric is intentionally baseline-relative: it measures behavioral preservation under masking, not absolute semantic correctness against reference outputs. Manual review of sampled Bad cases indicates that the heuristic is conservative in some cases, because acceptable rewrites can be classified as Bad when they diverge substantially from the baseline string. The main source artifacts for the result tables are outputs/phase10f/exp_10_1_to_37_comparison.md, outputs/phase10f/frontier_manual_verdict.md, and outputs/phase10i/phase10i_summary.md.

For the auxiliary thinking-phase expert isolation, we use phase-labeled routing events rather than generation-quality labels. The phase-separated pipeline labels tokens as prompt, think, or answer, then aggregates selected expert frequency into layer-by-expert matrices for each phase. The resulting heatmaps are descriptive routing analyses; they do not by themselves prove causal importance.

**Token budget selection:** Early experiments used a large generation length (max_new_tokens = 5000). Diagnostic runs revealed that large token budgets exacerbated semantic drift under aggressive masking, allowing the model to hallucinate or leak reasoning-like text into the final output. Conversely, overly restrictive limits (max_new_tokens = 800 or stopping on <think> tags) prematurely truncated valid solutions. The chosen parameter (max_new_tokens = 1200) was the best tested constraint in this decode-control sweep: it reduced runaway drift without the severe truncation observed at lower caps.

## 3.3 Masking Methodology

**Expert ranking:** For each targeted MoE layer, we compute the L2 norm of each expert's feed-forward weight tensor. Experts are ranked by ascending magnitude; the bottom-*p*% are selected for masking.

**Masking mechanism:** Selected experts are excluded by forcing their router logits to a large negative value before top-k selection. This differs from directly zeroing weights: the weight tensors remain intact, but routing is redirected to select from the remaining experts. This non-destructive intervention enables controlled and reversible experiments without modifying checkpoints.

Two masking modes are evaluated:

- **Hard masking:** Complete exclusion. Selected experts should receive zero routing selections. This is the primary mode for all main experiments.
- **Soft masking:** A penalty (-2.0 or -5.0) is added to router logits, reducing but not eliminating selection probability. This is evaluated as an auxiliary branch.

**Disjoint buckets:** Configurations use disjoint pruning intervals where relevant, so that each level evaluates a unique parameter subset.

**Integrity verification:** A mandatory mechanical check confirms masked_selection_hits_total == 0 for hard-mask experiments before quality conclusions are drawn.

### 3.4 Layer Region Definitions

| Region | Layers | Count |
|---|---|---|
| Early | 0-9 | 10 |
| Middle | 10-29 | 20 |
| Late | 30-39 | 10 |
| Very Late | 35-39 | 5 |

## 4. Results

### 4.1 Routing Characterization: Evidence for Depth-Dependent Redundancy

Before masking experiments, we instrumented the model's routing behavior across 878 prompts, collecting 208.5 million router event records. Normalized per-token routing weight matrices reveal a clear depth-dependent pattern:

- **Early layers (0-9):** Diffuse, low-confidence routing with no strongly dominant expert. Top-1 weight for the most-routed expert at layer 0 is only 0.088.
- **Mid-depth layers (10-25):** Highest routing concentration and confidence. Peak specialization occurs near layer 20, where the top-1 minus top-2 margin reaches approximately 0.024.
- **Late layers (30-39):** Routing confidence collapses. The top-1 margin drops from approximately 0.024 at layer 20 to approximately 0.006 at layer 37. Router entropy rises to 5.30-5.35, approaching the theoretical maximum of ln(256), approximately 5.55.

***[Figure 1 placeholder]*** *Router confidence and entropy by layer depth. Late layers show collapsed margins and near-maximal entropy, indicating diffuse, low-confidence routing. Reference: outputs/phase8_interactive.html, Tab 11.*

This diffuse routing in late layers provides a mechanistic hypothesis for masking tolerance: when no single expert dominates a layer's routing decisions, removing low-magnitude experts may redirect tokens to functionally similar alternatives with limited disruption.

Specialization scoring also identified task-specific experts with strong affinity for particular output domains alongside generic infrastructure experts with near-zero specialization scores. These observations motivate future expert-level scoring beyond the layer-level analysis presented here.

### 4.2 Auxiliary Routing Probe: Reasoning vs. Generation Phases

We also ran an auxiliary reasoning-aware routing probe to ask whether intermediate reasoning and final answer generation use different expert subsets. This is the "thinking head isolation" component of the project, but more precisely it is **thinking-phase expert isolation**: the artifacts isolate MoE expert usage, not transformer attention heads.

This probe is **not part of the main Qwen3.6 masking result**. In Qwen3.6, explicit <think> segmentation was not reliable enough for a clean phase-separated routing study. The full phase-separated routing collection therefore used a related Qwen3.5-35B-A3B model where reasoning boundaries were recoverable. The E1c collection processed 878 XLCoST C++ to Python prompts, converted the traces to a 401M-row routing table, and aggregated them into prompt, think, answer, global, and think-minus-answer heatmaps. A follow-on E2 branch repeated the same three-phase routing decomposition across additional XLCoST language pairs.

The isolation procedure is:

1. Generate outputs with recoverable thinking boundaries.
2. Split the token stream into prompt, think, and answer phases.
3. Record the selected MoE experts for each token and layer.
4. Aggregate routing frequency into separate layer-by-expert matrices for each phase.
5. Compare think and answer matrices to identify think-heavy, answer-heavy, and shared experts.

The auxiliary probe showed three qualitative patterns:

- **Reasoning phase:** Routing during intermediate reasoning was broad, with low-to-moderate activation across much of the expert pool.
- **Output generation phase:** Final-answer routing was sparser, with more inactive experts than the reasoning phase.
- **Phase contrast:** Differential heatmaps isolated phase-skewed experts: some cells are think-heavy, others are answer-heavy, and many form a shared core. The result is not a clean split into two disjoint expert sets.

***[Figure 2 placeholder]*** *Auxiliary heatmaps comparing expert activation during reasoning and answering phases on the phase-separated Qwen3.5 probe. Reference: outputs/think_routing_qwen35/plots/e1c_layer_blocks.html and outputs/e2_routing/plots/e2_think_vs_answer_v2.html.*

These auxiliary observations matter for pruning because they warn against scoring experts only from monolithic generation traces. A globally low-use expert could still be phase-specific, and an apparently redundant late-layer expert could be answer-heavy rather than reasoning-heavy. However, the isolation remains descriptive: we did not perform causal single-expert ablations that prove the identified think-heavy experts are necessary for reasoning. Therefore, these observations motivate future phase-aware expert scoring, but they should not be interpreted as direct evidence for the Qwen3.6 masking frontier reported below.

### 4.3 Depth-Dependent Sensitivity

The core regional ablation isolates masking to three depth regions independently, applying 40% hard masking to each while leaving all other layers unperturbed. Results on the 100-prompt evaluation set are shown in Table 1.

**Table 1: Layer-region ablation (100 prompts, 40% hard masking per region)**

| Masked Region | Good | Similar | Bad | G+S | Valid Output |
|---|---|---|---|---|---|
| Early (0-9) | 28 | 36 | 36 | 64 | 92 |
| Middle (10-29) | 43 | 24 | 33 | 67 | 96 |
| **Late (30-39)** | **67** | **13** | **20** | **80** | **95** |
| All layers (30%, reference) | 30 | 33 | 37 | 63 | 85 |
| All layers (40%, reference) | 22 | 33 | 45 | 55 | 83 |

***[Figure 3 placeholder]*** *Bar chart of G+S retention by masked region. Late-layer masking at 40% outperforms early-layer, middle-layer, and all-layer configurations. Reference data: exp_10_7, exp_10_4, exp_10_12-exp_10_14.*

Late-layer masking at 40% achieves G+S = 80/100, exceeding all-layer masking at 30% (63/100) while concentrating the intervention in only 10 of 40 MoE layers. Manual review of the 20 Bad cases from late-layer masking found many conservative false positives or acceptable rewrites, though the automated metric is retained for consistency across runs.

Scaling to 300 prompts confirms the regional result: late-layer 40% masking achieves 249/300 G+S, while flat all-layer 30% masking achieves only 150/300.

### 4.4 Discovery Sweep: Late-Layer Policy Search at 300 Prompts

With the depth-dependent result established, we explored masking schedules within and around the late-layer region across a 300-prompt discovery sweep. Table 2 shows the strongest candidates and representative failures.

**Table 2: 300-prompt discovery sweep (ranked by G+S)**

| Rank | Configuration | Good | Similar | Bad | G+S | Masked Experts | Avg Sim |
|---|---|---|---|---|---|---|---|
| 1 | **Late ramp: 30-34 @ 35%, 35-39 @ 55%** | **198** | **57** | **45** | **255** | **1,145** | **0.91** |
| 2 | Tiered: late hard 30% + soft next 15% | 199 | 53 | 48 | 252 | 1,140 | 0.90 |
| 3 | Global late budget: weakest 1,020 | 199 | 53 | 48 | 252 | 1,020 | 0.90 |
| 4 | Very late 35-39 @ 50% | 203 | 47 | 50 | 250 | 640 | 0.89 |
| 5 | Late 30-39 @ 40% | 194 | 55 | 51 | 249 | 1,020 | 0.89 |
| 6 | Preserve L39; 30-38 @ 45% | 194 | 53 | 53 | 247 | 1,035 | 0.90 |
| 7 | Late 30-39 @ 45% | 189 | 56 | 55 | 245 | 1,150 | 0.90 |
| 8 | Global late budget: weakest 1,150 | 196 | 49 | 55 | 245 | 1,150 | 0.89 |
| 9 | Late ramp: 30-34 @ 30%, 35-39 @ 50% | 200 | 43 | 57 | 243 | 1,020 | 0.89 |
| 10 | Late 30-39 @ 52% | 181 | 61 | 58 | 242 | 1,330 | 0.88 |
| 11 | Late 30-39 @ 42% | 198 | 43 | 59 | 241 | 1,070 | 0.89 |
| 12 | Late 30-39 @ 50% | 186 | 55 | 59 | 241 | 1,280 | 0.88 |
| 13 | L20-29 @ 10% + L30-39 @ 45% | 177 | 62 | 61 | 239 | 1,400 | 0.89 |
| 14 | Alternating even late layers @ 60% | 167 | 69 | 64 | 236 | 765 | 0.87 |
| 15 | Very late 35-39 @ 60% | 190 | 46 | 64 | 236 | 765 | 0.87 |
| 16 | L20-29 @ 10% + L30-39 @ 40% | 185 | 51 | 64 | 236 | 1,270 | 0.88 |
| 17 | L10-19 @ 5%, L20-29 @ 10%, L30-39 @ 40% | 177 | 57 | 66 | 234 | 1,390 | 0.87 |
| 18 | Late 30-39 @ 48% | 175 | 51 | 74 | 226 | 1,220 | 0.86 |
| 19 | Grouped downward expansion | 132 | 81 | 87 | 213 | 4,080 | 0.84 |
| 20 | **Flat all-layer 30%** | **88** | **62** | **150** | **150** | **3,040** | **0.73** |

***[Figure 4 placeholder]*** *Quality vs. masked-expert scatter plot for the 300-prompt discovery sweep. The late ramp is the top-scoring policy, but very-late-only and global late-budget policies occupy attractive points on the frontier. Reference: outputs/phase10f/exp_10_1_to_37_comparison.md.*

The discovery sweep supports four conclusions.

First, late-focused schedules dominate flat all-layer masking. Flat 30% all-layer masking is mechanically valid (masked_selection_hits_total == 0) but behaviorally poor, producing 150 Bad outputs at 300-prompt scale.

Second, the top policies are close enough that they should be interpreted as a frontier rather than as a single statistically settled winner. The top five candidates span only six G+S points (249-255) while masking between 640 and 1,145 experts.

Third, very-late layers are especially compressible. Masking only layers 35-39 at 50% achieves 250/300 G+S while masking 640 experts, nearly matching the late ramp while masking 505 fewer experts.

Fourth, downward expansion below layer 30 is consistently harmful. Adding 10% masking to layers 20-29 reduces quality relative to late-only policies, and extending into layers 10-19 degrades further.

### 4.5 Held-Out Validation at 500 Prompts

To test whether the 300-prompt frontier generalizes, we ran a larger 500-prompt held-out validation over three hard-mask frontier policies, using an unmasked baseline from the same prompt slice.

**Table 3: 500-prompt held-out validation**

| Policy | Masked Experts | Good | Similar | Bad | G+S | Avg Sim |
|---|---|---|---|---|---|---|
| **Very late 35-39 @ 50%** | **640** | **343** | **76** | **81** | **419** | **0.897** |
| Global late budget: weakest 1,020 | 1,020 | 336 | 74 | 90 | 410 | 0.895 |
| Late ramp: 30-34 @ 35%, 35-39 @ 55% | 1,145 | 317 | 91 | 92 | 408 | 0.892 |

***[Figure 5 placeholder]*** *Held-out quality vs. masked-expert frontier. On the 500-prompt validation slice, the narrow very-late-only policy is best by both quality and masked-expert count among the tested hard-mask frontier candidates. Reference: outputs/phase10i/phase10i_summary.md.*

The 500-prompt validation changes the interpretation of the 300-prompt sweep. The late ramp remains a strong discovery result, but it does not remain the best validation candidate. Instead, the very-late-only policy (layers 35-39 @ 50%) achieves the highest held-out G+S score while masking substantially fewer experts. This strengthens the broader depth-aware hypothesis while weakening any claim that the 30-34/35-39 ramp is uniquely optimal.

The practical readout is therefore:

- **Best 300-prompt discovery score:** late ramp 30-34 @ 35%, 35-39 @ 55%.
- **Best 500-prompt held-out quality/masked-expert tradeoff:** very-late 35-39 @ 50%.
- **Best simple broader comparator:** global late budget of 1,020 experts.

### 4.6 Compression Quantification

The very-late 35-39 @ 50% policy masks 640 of 10,240 total experts, corresponding to 6.25% of all experts and 50% of experts in the final five MoE layers. The 300-prompt late ramp masks 1,145 experts, corresponding to 11.2% of all experts and 44.7% of experts in layers 30-39.

These numbers describe expert eligibility removal, not realized checkpoint compression. Under the current inference-time masking implementation, the model file size is unchanged because masked expert weights remain present. Physical weight surgery would be required to convert these masking policies into actual memory, storage, and deployment savings.

### 4.7 Routing Width Reduction as a Separate Compression Axis

Independent of expert masking, we evaluated reducing the number of active experts per token from 8 to 6.

**Table 4: Top-k routing width reduction (100 prompts)**

| Configuration | Good | Similar | Bad | G+S |
|---|---|---|---|---|
| Baseline/reference (top-k = 8) | - | - | - | 100/100 |
| Top-k = 6, no masking | 90 | 10 | 0 | 100 |
| Top-k = 6 + late 40% masking | 60 | 17 | 23 | 77 |

Top-k reduction from 8 to 6, applied without expert masking, achieves no Good+Similar loss on this 100-prompt slice and shows a large observed wall-clock reduction relative to the available top-k=8 reference run. This timing result should be treated as a probe rather than a controlled latency benchmark, because full latency characterization requires matched prompt counts, hosts, batch settings, and generation lengths. Compounding k=6 with aggressive late-layer masking introduces substantial quality regression, suggesting that active-expert reduction and expert masking are not simply additive at the tested aggressiveness level.

## 5. Discussion

### 5.1 Why Late Layers Tolerate Masking

The results are consistent with a depth-dependent view of MoE representations. Early and middle layers construct foundational semantic and syntactic representations, including structural patterns, control flow, and algorithmic logic. Perturbations in these layers propagate forward and affect all later computation.

Late layers operate on already processed representations closer to the output distribution. The routing evidence supports this interpretation: late-layer entropy approaches the theoretical maximum and routing margins collapse. When routing is diffuse, experts within a layer may be more interchangeable, making the removal of low-magnitude experts less disruptive.

### 5.2 From "Best Schedule" to a Frontier

The 300-prompt discovery sweep initially suggests that the late ramp is the best policy. The 500-prompt validation shows a more nuanced result: the narrow very-late-only policy is stronger on the held-out slice and masks fewer experts. This does not invalidate the late-ramp result; it reframes it as one point on a frontier.

The stronger claim is not that one fixed schedule is universally optimal. The stronger and better-supported claim is that **useful MoE masking policies should be depth-aware, and the safe region begins much later than a uniform global pruning schedule assumes**.

### 5.3 Qualitative Failure Modes of Flat Masking

Flat all-layer 30% masking performs poorly despite being mechanically valid (masked_selection_hits_total == 0). This demonstrates a critical finding: mechanically valid masking does not guarantee behaviorally safe generation.

Manual inspection of the degraded outputs (the "Bad" category) reveals that the failure is rarely limited to simple syntax errors. Instead, the model suffers from severe semantic drift and loss of task-fidelity.

Many outputs are syntactically valid but solve a nearby, incorrect interpretation of the prompt. A dominant failure pattern is "reasoning leakage," where the masked model loses the boundary between intermediate computation and final output, leaking reasoning tokens directly into the required code block. This suggests that flat masking across early and middle layers compounds representational damage, fundamentally altering the model's interpretation of the task and its structural formatting constraints.

### 5.4 Limitations

1. **Inference-time masking, not physical pruning.** All experiments use router-logit manipulation, not permanent weight removal. Physical pruning may produce different behavior and is required to realize actual memory savings.

2. **Single model and single task.** Findings are specific to Qwen3.6-35B-A3B on one code translation task from the XLCoST benchmark. Generalization to other MoE architectures, natural-language tasks, and additional evaluation domains requires further study.

3. **Baseline-relative scoring.** The primary metric measures preservation relative to the unmasked baseline generation, not absolute semantic correctness. This is appropriate for sensitivity analysis but insufficient for final output quality claims.

4. **Heuristic similarity scoring.** Good/Similar/Bad classification uses syntactic validity checks and normalized string similarity. Complementary metrics such as AST similarity, CodeBLEU [17], execution-based evaluation, and human review would strengthen the evaluation.

5. **No confidence intervals.** Each configuration is evaluated as a single run. The top 300-prompt candidates differ by only a few examples, so bootstrap confidence intervals are needed before claiming strict statistical superiority among frontier policies.

6. **No post-masking fine-tuning.** Quality measurements are zero-shot under masking. Some degradation could likely be recovered through targeted SFT or routing-aware training.

7. **Reasoning-like generation artifacts.** The model sometimes produces reasoning-like text even under output-only instructions. These artifacts inflate generation length and can interfere with output extraction, but explicit reasoning-phase segmentation was not reliable enough in Qwen3.6 to support the auxiliary phase-routing claims directly.

## 6. Conclusion and Future Work

This paper presented a systematic layer-wise sensitivity analysis of Qwen3.6-35B-A3B, spanning regional ablations, 300-prompt policy discovery, and 500-prompt held-out validation. The central empirical finding is that **MoE expert masking sensitivity is strongly depth-dependent**: early and middle layers are fragile, while late and very-late layers tolerate substantial expert exclusion.

The best 300-prompt discovery policy is a late ramp, masking layers 30-34 at 35% and layers 35-39 at 55%, with 255/300 Good+Similar outputs. However, held-out 500-prompt validation favors a narrower very-late policy, masking layers 35-39 at 50%, which achieves 419/500 Good+Similar while masking only 640 experts. The conclusion is therefore not a single universal pruning recipe, but a robust depth-aware principle: **the final MoE layers are the dominant near-term compression target, and flat all-layer masking is a poor strategy.**

Future work should proceed along the following paths:

1. **Physical weight surgery:** Convert the best hard-mask policies, especially very-late 35-39 @ 50% and global late budget 1,020, into actual checkpoint modifications to measure realized memory and deployment savings.
2. **Semantic evaluation:** Add AST similarity, CodeBLEU, execution-based tests where possible, and stratified manual review of Bad cases.
3. **Confidence intervals:** Bootstrap the 300- and 500-prompt comparisons so small differences among frontier policies are reported with uncertainty.
4. **Training-based recovery:** Fine-tune physically pruned or hard-masked models to test whether remaining experts can recover lost behavior.
5. **Dynamic active-expert reduction:** Expand on the top-k = 6 probe with matched latency benchmarks and lower-aggression combinations with very-late masking.
6. **Phase-aware expert scoring:** Compare magnitude ranking against routing frequency, routing weight, task affinity, and the auxiliary prompt/think/answer isolation signals.
7. **Cross-task generalization:** Repeat the depth-sensitivity protocol on additional tasks and evaluation domains to determine whether very-late compressibility is architecture-general or task-specific.